# Trajectory Flow Map: Graph-Based Approach to Analysing Temporal Evolution of Aggregated Traffic Flows in Large-Scale Urban Networks

**Jiwon Kim, Corresponding Author**
The University of Queensland
Brisbane St Lucia, QLD 4072, Australia
Email: jiwon.kim@uq.edu.au

**Kai Zheng**
The University of Queensland
Brisbane St Lucia, QLD 4072, Australia
Email: kevinz@itee.uq.edu.au

**Jonathan Corcoran**
The University of Queensland
Brisbane St Lucia, QLD 4072, Australia
Email: jj.corcoran@uq.edu.au

**Sanghyung Ahn**
The University of Queensland
Brisbane St Lucia, QLD 4072, Australia
Email: sanghyung.ahn@uq.edu.au

**Marty Papamanolis**
The University of Queensland
Brisbane St Lucia, QLD 4072, Australia
Email: s.papamanolis@uq.edu.au

# Trajectory Flow Map: Graph-based Approach to Analysing Temporal Evolution of Aggregated Traffic Flows in Large-scale Urban Networks

## ABSTRACT

This paper proposes a graph-based approach to representing spatio-temporal trajectory data that allows an effective visualization and characterization of city-wide traffic dynamics. With the advance of sensor, mobile, and Internet of Things (IoT) technologies, vehicle and passenger trajectories are being increasingly collected on a massive scale and are becoming a critical source of insight into traffic pattern and traveller behaviour. To leverage such trajectory data to better understand traffic dynamics in a large-scale urban network, this study develops a trajectory-based network traffic analysis method that converts individual trajectory data into a sequence of graphs that evolve over time (known as dynamic graphs or time-evolving graphs) and analyses network-wide traffic patterns in terms of a compact and informative graph-representation of aggregated traffic flows. First, we partition the entire network into a set of cells based on the spatial distribution of data points in individual trajectories, where the cells represent spatial regions between which aggregated traffic flows can be measured. Next, dynamic flows of moving objects are represented as a time-evolving graph, where regions are graph vertices and flows between them are treated as weighted directed edges. Given a fixed set of vertices, edges can be inserted or removed at every time step depending on the presence of traffic flows between two regions at a given time window. Once a dynamic graph is built, we apply graph mining algorithms to detect change-points in time, which represent time points where the graph exhibits significant changes in its overall structure and, thus, correspond to change-points in city-wide mobility pattern throughout the day (e.g., global transition points between peak and off-peak periods).

## INTRODUCTION

Recent years have witnessed a growing interest in an aggregated modelling of traffic networks, i.e., modelling and analysis approaches that investigate aggregated properties of network-wide traffic dynamics, aiming at describing collective phenomena of the underlying systems and predicting their behaviours in a more robust manner. In the field of traffic flow theory, one of such approaches is the study of the Macroscopic Fundamental Diagram (MFD), which aims to characterise the network-wide traffic flow behaviours using network-level flow-density relationships *(1–4)*. There are also efforts for modelling network-wide travel time reliability characteristics by studying the relationship between travel time variability and overall congestion level *(5, 6)*. With the availability of large-scale trajectory data, much attention has been given to the application of data science and data mining techniques to better analyse spatio-temporal evolution of traffic flows at the network-scale, leading to many exciting developments from various research fields. Examples include the works detecting major traffic streams using trajectory clustering *(7)*, discovering causal interactions between regions *(8)*, and visual analysis for mass mobility dynamics *(9)*. This study has been also motivated by the availability of trajectory data that are being increasingly collected on a massive scale around the city. Trajectory data provide detailed information on how the given network is being utilized by vehicles and travellers over space and time, offering valuable insight into traffic patterns and traveller behaviours. Yet, in many cases, trajectory data are still treated in the same way as traditional traffic detector data and researchers and practitioners are not fully leveraging unique benefits of trajectory data for their analysis. This study aims to fill this gap by proposing a graph-based analysis framework for mining, characterising, and visualising traffic dynamics in a city, which has a particular focus on leveraging trajectories of moving objects to assist in an aggregated modelling of network-wide traffic dynamics.

A graph-based representation provides a powerful means of describing complex relationships among entities in a dataset in a simple and intuitive way, and studying the properties of these graphs provides useful information about the internal structure of the dataset. In many real-world systems, relationships between entities evolve over time and such time-varying relationships or connections can be represented as a sequence or time-series of graphs, namely a dynamic graph or time-evolving graph. Traffic flows in a road network can be naturally represented as a dynamic graph, where different parts of the network are represented as graph vertices and traffic flows between them are represented as directed edges with time-varying weights. Unlike the traditional road network graph, where vertices and edges represent intersections and links, respectively, and network connections represent static routes, this "flow graph" or "flow map" connects geographically distributed areas through dynamic flows of vehicles or people and is constantly undergoing changes to its network connection structure *(9)*. For instance, given a fixed set of vertices, which represent a pre-defined set of regions of interest, edges can be inserted or removed at every time step depending on the presence of traffic flows between two regions at a given time window. The recent availability of area-wide trajectory data such as GPS vehicle trajectories or smart card passenger trajectories has enabled the construction of such dynamic flow graphs *(7)*.

This study proposes a graph-based analysis framework that converts individual trajectory data into a compact and informative representation of dynamic graphs and explores new ways of measuring, visualizing, and characterizing network-wide traffic dynamics and mobility patterns. A particular application addressed in this paper is to create a robust profile of daily network traffic patterns using a graph summarization method applied to a *graph series* or a sequence of aggregated flow graphs throughout the day. FIGURE 1 (a) illustrates an example of daily graph series that is constructed from bus passenger trajectories obtained in Brisbane, Australia. The entire day is divided into a sequence of time slices with a fixed time window length (30 minutes in this example) and a graph of aggregated flows is constructed for each time slice, which provides a snapshot of network-wide traffic flow pattern or spatial distribution of flows for that particular time window. These snapshots are then connected in a sequence in the order of time, providing the information on the temporal evolution of the spatial traffic distribution patterns. Once we obtain such a graph series of daily traffic, we can apply a change point detection technique to identify time points where the graph structure changes significantly compared to their previous time slices and segment the daily graph series into a set of graph segments that can be used to characterise and summarise overall traffic patterns in different time-of-day periods.

**RELATED WORK**

Time-evolving graph (a.k.a. dynamic graph) is usually modelled as a sequence of static graphs, each representing the status of the network at particular time instance. The main advantage of dynamic graphs over static graphs lies in their power to encode transition or change with time by allowing insertion, deletion, and update of vertices and edges between consecutive graph snapshots. In recent years, there are many works on evolving graph problem. Sun et al. *(10)* present dynamic tensor analysis, which incrementally summarizes tensor streams (high-order graph streams) as smaller core tensor streams and projection matrices. Sun et al. *(11)* propose GraphScope which discovers communities in large and dynamic graphs, as well as detects the changing time of communities. Tong et al. *(12)* propose a family of Colibri methods to track low rank approximation efficiently over time. The authors in *(13–15)* propose distinct measure functions such as maximum common subgraph to detect when the graph changes. However, these measure functions incur extensive computational cost and thus cannot be used to monitor the changes over entire networks.

There is an increasing interest in mining dynamic graphs. Borgwardt et al. *(16)* apply frequent-subgraph mining algorithms to time series of graphs, and extract subgraphs that are frequent within the set of graphs. Bifet and Gavald in *(17)* present three closed tree mining algorithms to mining frequent closed tree in evolving graphs. The extraction of periodic or near periodic subgraphs is considered in *(18)* where the problem is shown to be polynomial. Inokuchi and Washio in *(19)* and Robardet in *(20)* discuss finding frequent evolving patterns in dynamic networks. Chan et al. in *(21)* introduce a new pattern to be dis- covered from evolving graphs, namely regions of the graph that are evolving in a correlated manner. All of these works are to finding a subgraph sequence pattern, such that (a) its embedding in a graph sequence is frequent and (b) the behaviour of these embedding are identical over time. Liu et al. in *(22)* proposes a random walk model with restart to discover subgraphs that exhibit significant changes in evolving networks. Significance is measured by the total change of similarities between vertex pairs inside subgraphs. After finding the most significant changing vertices, a clustering-manner algorithm is used to connect these vertices into subgraphs. However, this method is only concerned with changes between two graphs, and cannot quantify changes of subgraph in a time interval. Yang et al. *(23)* studied finding a subgraph which change most in an evolving graph, based on the cumulated connectivity change that is shown to be effective in identifying the most changed subgraph with a small number of unchanged edges included

Ren et al. *(24)* propose the FVF framework to answer shortest path query for all snapshots in evolving graphs. Aggarwal et al. *(25)* use a structural connectivity model to detect outliers which act abnormally in some snapshots of graph streams. Feigenbaum et al. *(26)* explore the problem related to compute graph distances in a data-stream model, whose goal is to design algorithms to process the edges of a graph in an arbitrary order given only a limited amount of memory. Tantipathananandh et al. *(27, 28)* propose frameworks and algorithms for identifying communities in social networks that change over time. Kumar et al. *(29)* aimed to discover community bursts in a time-evolving blog graph. Their algorithm first extracts dense subgraphs from the blog graph to form potential communities. Then, the bursts within all the potential communities are identified by modelling the generation of events by an automaton. Bansal et al. *(30)* focused on seeking stable keyword clusters in a keyword graph, which evolves with additions of blog posts over time.

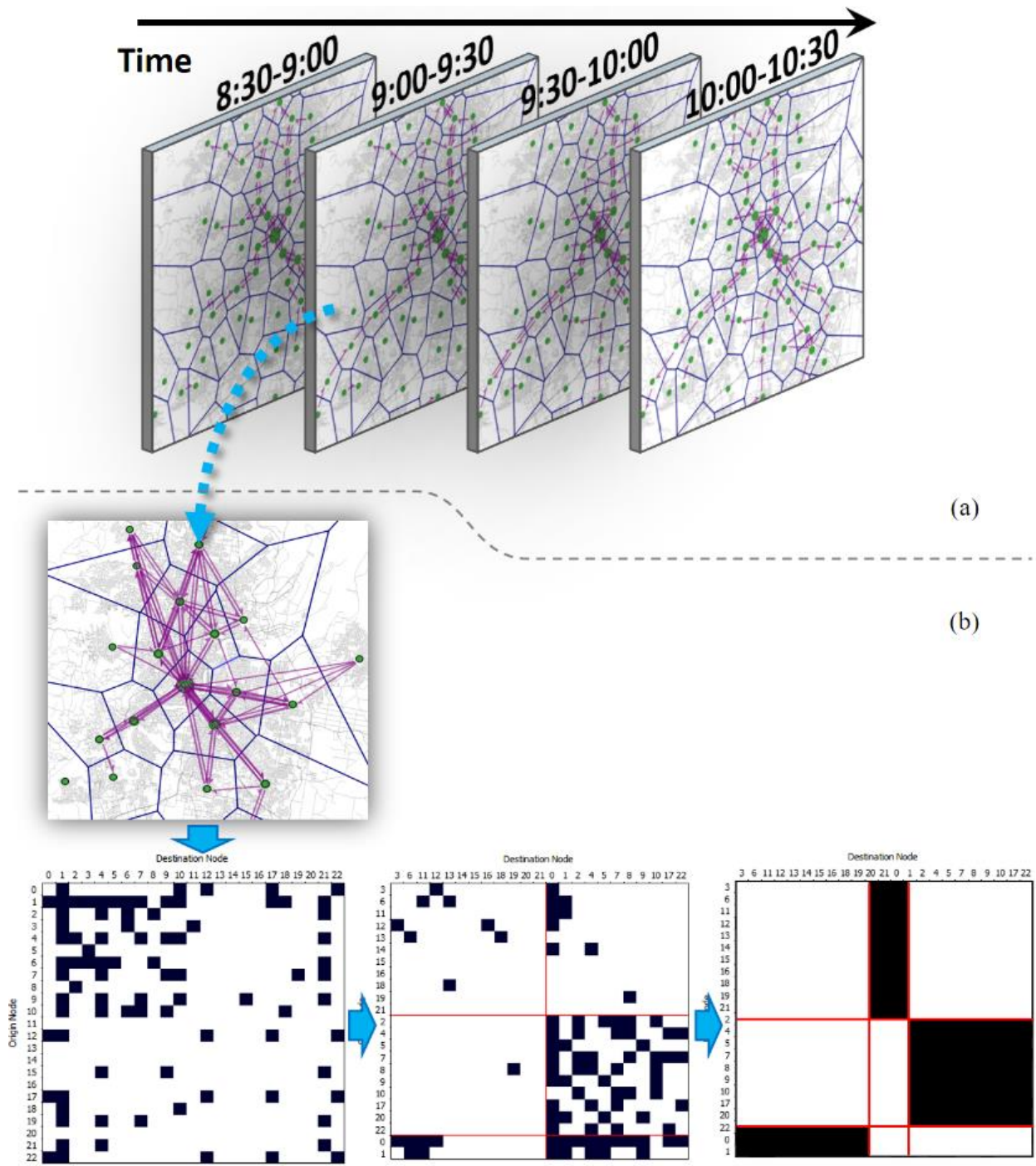

**FIGURE 1 Illustration of a trajectory flow map, (a) a dynamic graph of aggregated traffic flows constructed from trajectory data and (b) an approach to compressing and summarising each graph snapshot using its adjacency matrix.**

## METHODOLOGY

In this section, we describe detailed methodology and techniques for characterising spatial and temporal patterns of network-wide traffic flows using a *trajectory flow graph*. The term "trajectory flow graph" is used to represent a graph of *aggregated flows* built from *trajectory* data. The proposed framework consists of three stages: (i) network partitioning, (ii) graph generation, and (iii) change point detection. Detailed procedures for each stage are discussed below.

### Partitioning a Network into Cells

A trajectory of moving object is a time-ordered sequence of points $p(x, y, t)$ that represent the x and y coordinates of the moving object at time $t$. In order to convert trajectory data into a dynamic graph of aggregated flows, the first step is to define spatial regions for which aggregated traffic flow will be computed. Below we describe an approach to partitioning the entire network into cells with similar size by clustering the data points in trajectories. Alternatively, one could use suburb or postal area boundaries to define spatial regions for the traffic flow analysis. The network partitioning method adopted in this study is based on the method developed by Adrienko and Adrienko *(31)*, where trajectory data points are grouped in space based on a desired spatial radius. Then, each point group is represented by a cell or partition in the network, where the cell boundaries are determined by a Voronoi tessellation.

Given a set of trajectories, we first extract "seed points" that will be used in constructing point groups or cells. One could use all the data points in the trajectory dataset or select only a subset of data points. Depending on the choice of seed points, the resulting network partitioning may have different implications. For instance, if we use trajectories' origin and destination points only, the point groups will tend to be cantered at major origin and destination points and, thus, the network will end up being partitioned into a set of representative origin and destination regions. Let $S$ be the set of pre-defined seed points and $\gamma$ represents a parameter specifying the desired radius of a cell. The procedure for grouping seed points are described as follows:

*Procedure 1: Grouping points into cells with a desired radius*
**Input:** *set of seed points S, desired cell radius γ*
**Output:** *set of cells C*
  1.  *Initialize C = { }.*
  2.  *For each seed point s in S,*
        a.  *Find the closest cell c in C, where the distance between s and c's centroid is less than or equal to γ and c's centroid is closer to s than any other cell centroids in C.*
        b.  *If such c is not found, create a new cell c and assign s to c. Add c to C.*
        c.  *Assign s to c and update c's centroid as the mean point of all member points in c.*
  3.  *For each cell c in C, remove all member points while keeping its centroid.*
  4.  *For each point s in S, find the closest cell c in C and redistribute s to c.*

Once all the seed points are grouped into cells and the centroid of each cell is obtained based on the mean position of the seed points assigned to the cell, the cell boundaries can be determined by building a Voronoi diagram, using those cell centroids as Voronoi generator points *(31, 32)*. A Voronoi diagram partitions a plane into convex polygons such that each polygon contains exactly one generator point and all locations in a Voronoi polygon are close to its own generator point than any other cells' generator points in the diagram. Correspondingly, the points on the boundary between two Voronoi polygons are equidistant to their two generator points. Using the Voronoi diagram constructed from the cell centroids, we partition the network into spatial cells. FIGURE 2 (a)-(b) illustrate this procedure. FIGURE 2 (a) shows original trajectory data displayed on the region map. The trajectory data used in this example include 1000 bus passenger trajectories obtained in Brisbane, Australia. Points in trajectories represent bus stops and the origin point of each trajectory is shown as green circle in the figure. In FIGURE 2 (b), all the points in the original trajectories are divided into cells, where different point colours represent different point clusters. The centroids of point groups are shown as black circles which are then used as generator points for constructing Voronoi polygons shown in thick lines.

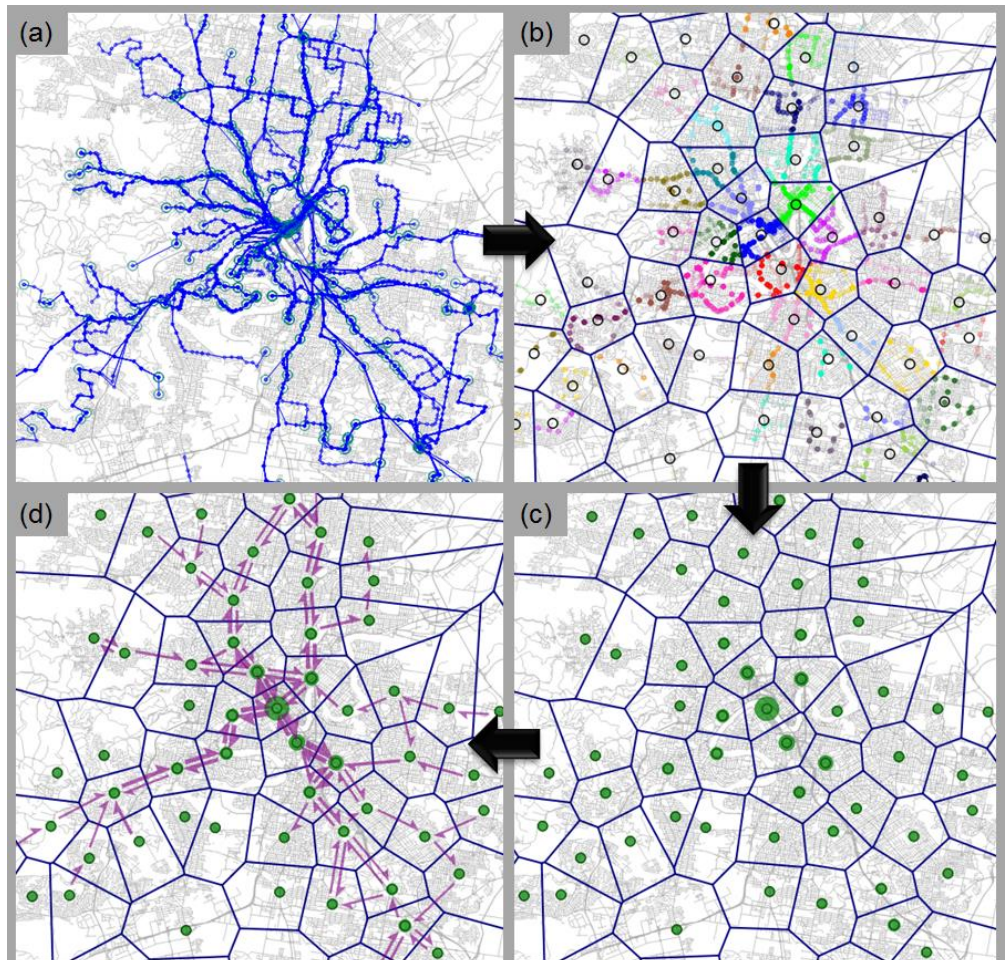

**FIGURE 2 Overall procedure of converting trajectory data into a dynamic graph of aggregated flows: (a) 1000 bus passenger trajectories, (b) partition the network into cells based on the data points in trajectories (c) set a graph vertex at each cell representing a geographical region; (d) connect vertices with bi-directional edges based on traffic flows between two regions.**

**Generating a Dynamic Graph of Aggregated Flows**

Once the network is partitioned into regions, these regions are used to define the vertex set of a trajectory flow graph that we will build in the second stage of the framework. In order to build a dynamic graph, we divide a day into $T$ time slices with a fixed length $\Delta t$. For each time slice $t = 1, ..., T$, we then build a trajectory flow graph, $G_t = (V, E_t \subseteq (V \times V))$, where $V$ is the vertex set that represents the spatial cells or regions identified above and $E_t$ is the edge set at time slice $t$ that represent traffic flows between cells. A time-ordered sequence of graph snapshots forms a trajectory flow graph series, denoted by $G = \{G_t\}_{t=1}^{T}$. It is noted that the vertex set $V$ is not time dependent since we assume that the spatial regions of interest are fixed as we are only interested in characterising time-varying traffic flow patterns across those cells that are already defined. This is to provide a common ground for comparing graphs across different times of day as well as across different days. For each time slice, we identify portions of trajectories (or sub-trajectories) that pass each cell during that time slice and obtain the information on the aggregated trajectory flows between each pair of cells as well as a sequence of cells visited by each trajectory.

*Defining Graph Edges*

In the trajectory flow graph, edges are the ones that define the dynamic characteristics of the graph. Different rules for adding edges provide different representations of the underling trajectory data. In this study, an edge is added from region $i$ to region $j$ if there is a trajectory visiting these two regions in that order, but not necessarily consecutively, during a given time window. Regions $i$ and $j$ are not necessarily the trajectory's origin and destination, respectively. For instance, if a trajectory passes three regions A, B, and C in this sequence, three edges can be added by connecting all region pairs that appear in the trajectory's visit sequence: A→B, A→C, and B→C. In this graph, edges reflect the existence of interactions between regions during a particular time interval, providing information on region connectivity and reachability through actual traveller flows.

We further apply a filter based on the speed of the region-to-region traffic to characterise the quality of the connections. Let $TR_{i,j}(t)$ denote a set of sub-trajectories, extracted from the original trajectory set, that start from region $i$ and end in region $j$ during time slice $t$ regardless of their intermediate regions visited between $i$ and $j$. The average speed of this region-to-region traffic $TR_{i,j}(t)$ during time slice $t$, denoted by $v_{i,j}(t)$, can be computed based on the Edie's definitions *(33)* as follows:

$$v_{i,j}(t) = \frac{\sum_{k \in TR_{i,j}(t)} d_k}{\sum_{k \in TR_{i,j}(t)} \tau_k} \tag{1}$$

where $\boldsymbol{d_k}$ is the travel distance covered by sub-trajectory $k$ and $\boldsymbol{\tau_k}$ is the travel time spent by sub-trajectory $k$. Then, for a given speed threshold $\boldsymbol{v_c}$, an edge is added or removed between two regions by comparing region-to-region traffic speed $v_{i,j}(t)$ with $\boldsymbol{v_c}$. Two methods can be possible: Method 1, where edges are added when the traffic speed is lower than $\boldsymbol{v_c}$, and Method 2, where edges are added when the traffic speed is higher than $\boldsymbol{v_c}$, as shown in Eqs. (2) and (3) below, respectively. The former graph—*a graph of low-speed flow*—will represent region-to-region connectivity with low quality traffic while the latter graph—*a graph of high-speed flow*—will represent region-to-region connectivity with high quality traffic. Both methods are tested in this study.

*Method 1: Graph of low-speed flow*

$$e_{i,j}(t) = \begin{cases} 1, & v_{i,j}(t) \le v_c \\ 0, & v_{i,j}(t) > v_c \end{cases} \tag{2}$$

*Method 2: Graph of high-speed flow*

$$e_{i,j}(t) = \begin{cases} 1, & v_{i,j}(t) > v_c \\ 0, & v_{i,j}(t) \le v_c \end{cases} \tag{3}$$

where $e_{i,j}(t)$ represents the state of the edge from region $i$ and region $j$ at time slice t (1 if an edge exists, 0 otherwise).

**Detecting Change Points in Time-evolving Trajectory Flow Graph**

One of the benefits of using a graph-based representation is that we can capture a more abstract view of the underlying data by applying various graph compression and summarization methods. In this paper, we use a graph partitioning technique to compress the trajectory flow graph for each time slice and attach similar time slices together to segment the entire daily graph series into a small number of graph segments, which allow us to exploit patterns and regularity in city-wide traffic flows during different time periods of the day. The third stage of the proposed framework is thus to create a temporal profile of city-wide traffic flows by segmenting a daily trajectory flow graph series. In segmenting a graph series, we apply a parameter-free change detection approach called *GraphScope (11)*, which uses the *Minimum Description Language* (MDL) principle to automatically detect natural clusters of graph vertices in a graph snapshot and find "change points"—points in time where the entire structure of the graph changes—in a sequence of graph snapshots. In this approach, each graph snapshot $G_t$ is expressed as a $|V| \times |V|$ binary adjacency matrix, where rows represent upstream regions and columns represent downstream, as illustrated in FIGURE 1 (b). Entry $(i, j)$ of the adjacency matrix is 1 if there is an edge from region $i$ to region $j$ and 0 otherwise. By reordering and grouping rows and columns, one can aim to divide the matrix entries into rectangular regions or blocks of ones and zeros as homogenous as possible, where a block of ones represents a fully interconnected vertex group and a block of zeros represents a group of vertices with no connection between them. Such a rearranged adjacency matrix will allow us to encode the entire graph structure in terms a few blocks of ones and zeros thereby reducing the cost of compressing the graph data. As such, this matrix rearrangement or graph partitioning can be achieved by minimizing the graph encoding cost measured by an MDL-based cost function, which is based on Shannon entropy. After finding the best row and column partitions in the adjacency matrix for each time slice, consecutive time slices that are similar in terms of their partition structures are grouped together such that the total encoding cost for the entire graph series can be minimised. For more details about the MDL-based graph partitioning and segmentation, readers are referred to the relevant literature *(11, 34)*.

Given graph series $G = \{G_t\}_{t=1}^{T}$, the outcome of this procedure is a set of change points $T^c$ in the graph series and a set of graph segments $G = \left\{G_{(s)}\right\}_{s=1}^{|T^c|+1}$, where $G_{(s)}$ denotes the $s^{th}$ segment in the graph series representing a consecutive sequence of graph snapshots between the $(s\text{-}1)^{th}$ and $s^{th}$ change points. Each graph segment has its associated compressed adjacency matrix that describes the underlying graph connection structure for all the time slices belonging to that segment. Using these change points and identified graph segments, we can create a concise profile of the temporal evolution patterns in a network-wide traffic flow, which will be discussed in more detail in the case study below.

## CASE STUDY
### Study Area and Data
The trajectory data used in this study are bus passenger trajectories generated from public transit smart card data in Brisbane, Australia. The smart card system in Brisbane (terms *go* card) is the major means by which over 80% of all urban public transport passengers pay transit fares. In the *go* card system, users need to touch their *go* cards to the on-board readers at the time of both boarding and alighting. Hence, the Brisbane smart card database includes information about both boarding and alighting locations and times. The *go* card data have been supplied by TransLink, the transit agency in Brisbane, Australia. By matching the *go* card transaction records with the underlying bus route information obtained from the associated General Transit Feed Specification (GTFS) data, we generate bus journey trajectories of individual bus passengers, where each trajectory represents a sequence of bus stops along the full journey of a particular passenger including transfers between his/her origin and final destination, with the location and time information for each stop.

The proposed framework is demonstrated using passenger trajectories from 12 March 2013, which was a normal Tuesday. A total of 66,800 trajectories are obtained from the time period between 5:00am and 11:59pm. We only included trajectories longer than 3km to reduce noise due to very short trips. All the data points in trajectories are used as seed points $S$ for building spatial cells. For the spatial and temporal resolutions of the graph, this study uses desired cell radius $\gamma = 3$ km and time slice width $\Delta t = 60$ min, respectively.

### Results and Analysis
*Finding Speed Threshold for Change Point Detection*
After conducting Procedure 1, the network was partitioned into 57 regions as shown in FIGURE 3 (a). FIGURE 3 (b) and (c) present graph snapshots from time slices 7:00-8:00 and 10:00-11:00, respectively, where edges show entire region-to-region flows without imposing any speed threshold criterion. The edge connection structure appears similar in these two snapshots because any two regions in this area could be reachable within a 1-hour window, given that there exist passenger demands in those regions. Once we apply speed criterion $v_{i,j}(t) \leq 20$km, however, the graph structure becomes very different between two time periods. Many edges are still present during 7:00-8:00 indicating that the average region-to-region travel speed is lower than 20km in many region pairs, while only a few edges appear during 10:00-11:00 indicating that only a few region pairs experience the average connection speed lower than 20km. This suggests that changes in the graph structure over different time periods could be more effectively detected by selecting the speed criterion that can reveal hidden differences.

In order to understand the effect of speed criterion on graph change point detection results, we further investigated the number of edges in each graph snapshot constructed under different edge generation methods, where two methods Eqs. (2) and (3) with varying levels of speed threshold values ($v_c$) were compared. The total number of edges in a graph snapshot is used as a simple aggregated measure to quantify the similarity between two graph snapshots. The results are provided in FIGURE 4, where the left column shows the results associated with Method 1 in Eq. (2) and the right column shows the results associated with Method 2 in Eq. (3). In the first row (a, b), plots show the number of edges in a graph snapshot for different time slices by different speed thresholds. Overall, the number of edges varies between 0 and 500 (per time slice). When the criterion is too loose (e.g., ≤45km or >15km), nearly all possible regions remain connected throughout the day. Conversely, when the criterion is too tight (e.g., ≤15km or >45km), the number of edges is constantly very small. It appears that $v_c$ with the range between 20 and 30km/h is suitable for producing a noticeable level of variations in the number of edges across time-of-day periods. In the second row (c, d), the number of edges is divided by the maximum edge count (per time slice) within each graph series to show the relative changes in the number of edges across time slices under the same speed criterion. Among the criteria with $v_c = [20, 30]$, ≤20km in figure (c) and >30km in figure (d) lead to the largest variations in their relative edge count. The third and fourth rows represent the same information in the first and second rows using speed threshold level as an x-axis, respectively. It can be easily observed from figures (e) and (f) that the number of edges dramatically increases or decreases during a relatively short range of $v_c = [20, 30]$. In figures (g) and (h), one can observe, for a given x value (edge criterion), the range of edge connections that the associated graph series exhibits throughout the day from the distribution of time-of-day curves over the y-axis. In terms of relative edge count, ≤20km and >30km shows the highest variations across different time slices.

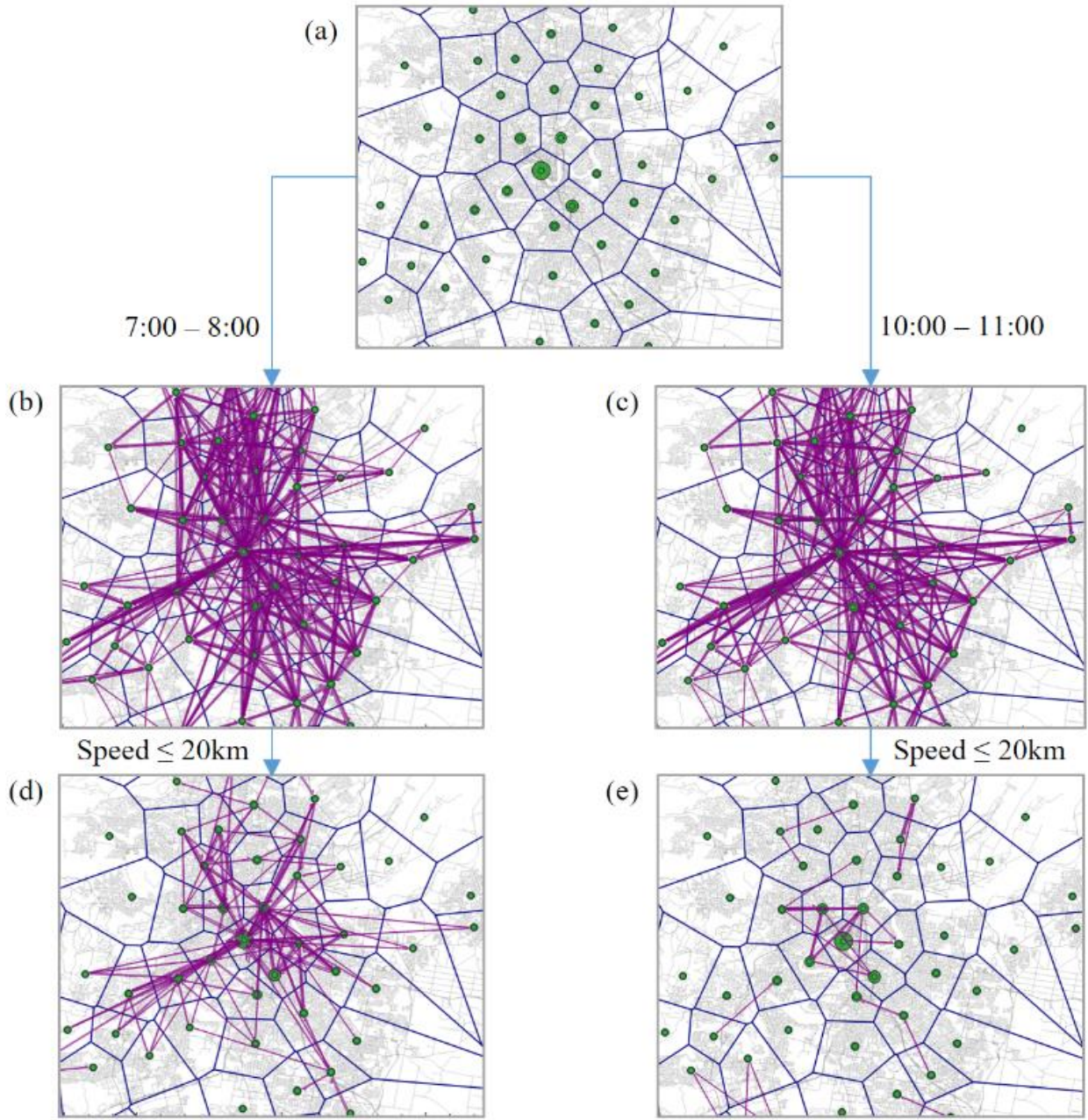

**FIGURE 3 Network partitioning and graph generation results.**

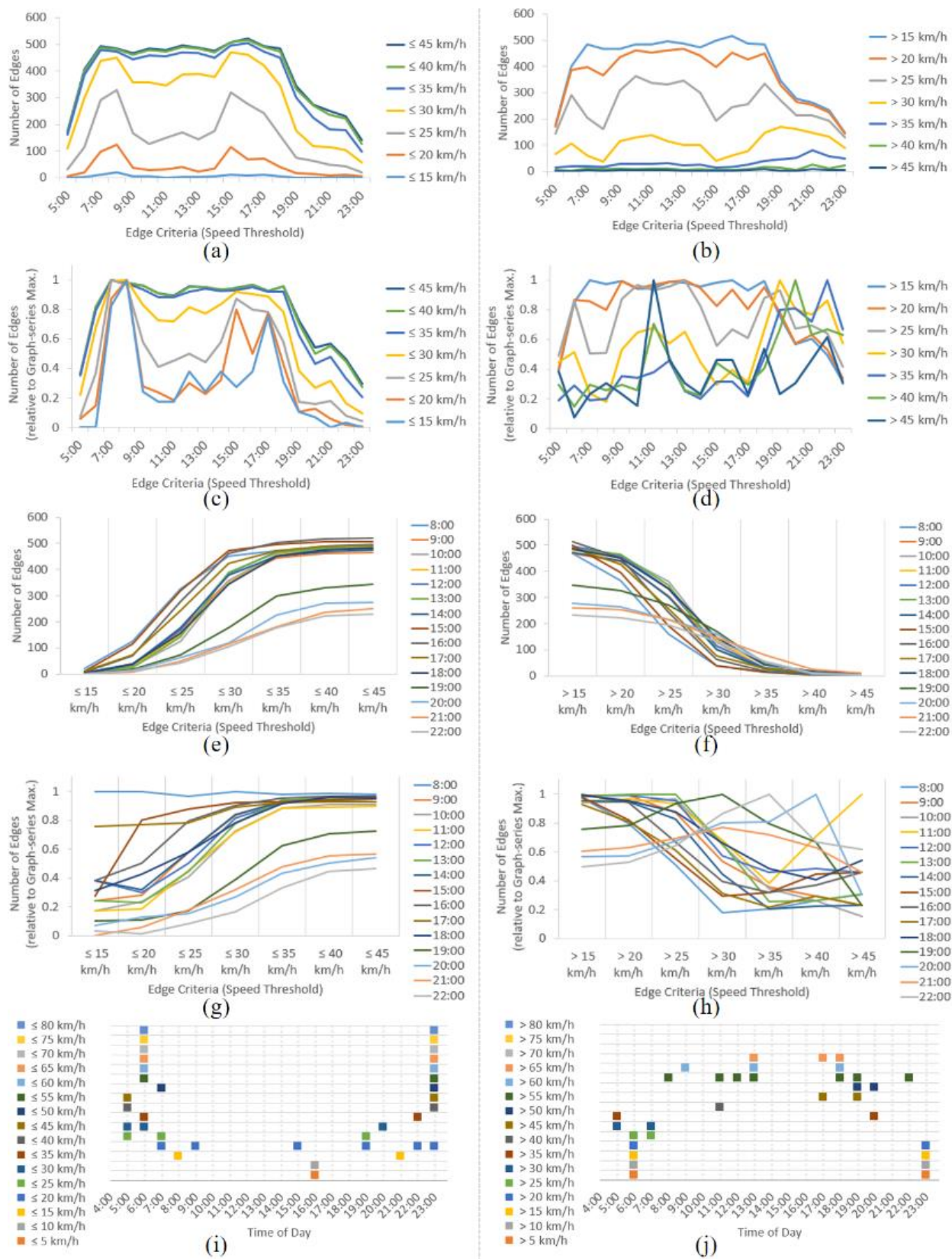

**FIGURE 4 Effects of speed threshold criterion on the structure of generated graph.**

*Effects of Speed Threshold on Change Point Detection Results*
Next, we performed the actual process of change point detection. With the time slice width of 60min, we have a total of 19 graph snapshots that form a trajectory flow graph series, i.e., $G = \{G_t\}_{t=1}^{19}$. The graph series $G$ is then segmented based on the change points detected by the GraphScope algorithm described above. We applied the algorithm to a total of 32 graph series constructed under different edge generation criteria. For each of two methods in Eqs. (2) and (3), 16 speed threshold ($v_c$) values were tested ranging from 5km/h to 80km/h with a 5km/h increment. The results are summarized in FIGURE 4 (i) and (j), where y-axis represents graph series with different criteria and x-axis represents different time slices. Square markers on the chart indicate the change points detected for each graph series. The results are somewhat consistent with the observations made with the number of edges above. When the criterion is loose (e.g., ≤45-80km or >5-15km), there is no change point detected during the day except for two points at the beginning and end of the day because the graphs are always heavily connected. When the criterion is tight (e.g., ≤5-15km or >45-80km), Method 1 and Method 2 show slightly different results. Fewer change points are detected under Method 1 in this range in FIGURE 4 (i), while there are even more change points are detected under Method 2 in FIGURE 4 (j) such as 7 points under >55km/h. It is found that this is because a set of edges meeting >45-80km (i.e., speed higher than a high-speed criterion) are more random and irregular, which vary widely from time slice to time slice, whereas a set of edges meeting ≤5-15km (i.e., speed lower than a low speed-criterion) are more consistent and systematic, which are always found around major commuting regions. As such, in the graph series from >45-80km, the GraphScope algorithm tends to detect a change point whenever such irregular connections appear during the day where graph snapshots are mostly empty due to the tight criterion.

Among the criteria with $v_c = [20, 30]$, where a decent number of edges and their variations are observed (as discussed above), Method 1 (≤20-30km) seems to provide better results than Method 2 (>20-30km) as the former identifies 3-6 change points in the middle of the day, whereas the latter does not detect any change points except for the beginning and end of the day. Especially, the criterion ≤ 20km detects 6 changes points throughout the day. To understand how and what changes have been detected, we take a closer look at the case of ≤ 20km in the next section.

*Investigating Changes in Time-evolving Graph*
FIGURE 5 shows the change point detection results for the graph series with ≤ 20km. FIGURE 5 (a) depicts how the graph series with 19 time slices are divided into 7 graph segments, i.e., $G = \{G_{(s)}\}_{s=1}^{7}$, where the y-axis represents the final (minimized) graph encoding cost for each identified graph segment, in the unit of cost per hour. We can observe that different segments incur different levels of encoding costs. The more complex the graph structure is, the higher the cost will be. FIGURE 5 (b)-(c) show the adjacency matrices of the 2nd and 4th graph segments, where the rows and columns of each matrix has been reordered such that the matrix entries can be partitioned into row groups (RG) and column groups (CG) to form a few homogenous blocks. The partitioned adjacency matrix for each graph segment then captures the underlying graph connection structure for all the time slices associated with that segment. We can see that there is a clear difference in the adjacency matrix between two consecutive graph segments, suggesting that the tested change point detection method was able to detect points in time where the graph structure changes significantly. The structure of $G_{(2)}$, which represents the traffic during a morning peak period (6:00-9:00), is the most complex, followed by $G_{(4)}$, which represents the afternoon peak period (15:00-19:00). FIGURE 5 (d)-(g) show a group of regions and their connections for selected blocks in the adjacency matrix. For $G_{(2)}$, regions associated with the block of RG2 and CG1 are shown in (d); RG1 and CG3 in (e); and RG4 and CG1 in (f), respectively. These partitions have the shape of a vertical line indicating *many-to-one* connections, where edges are coming from many upstream regions and going to a single downstream region. This captures the inbound traffic pattern that is typical during the morning commuting period in Brisbane, where the single downstream region corresponds to the Brisbane central business district (CBD). For $G_{(4)}$, the block of RG1 and CG1 is shown in (g). The block now shows a horizontal line indicating *one-to-many* connections, where edges are coming from a single upstream region and going to many downstream regions. This captures the outbound traffic pattern that is typical during the afternoon peak period in Brisbane.

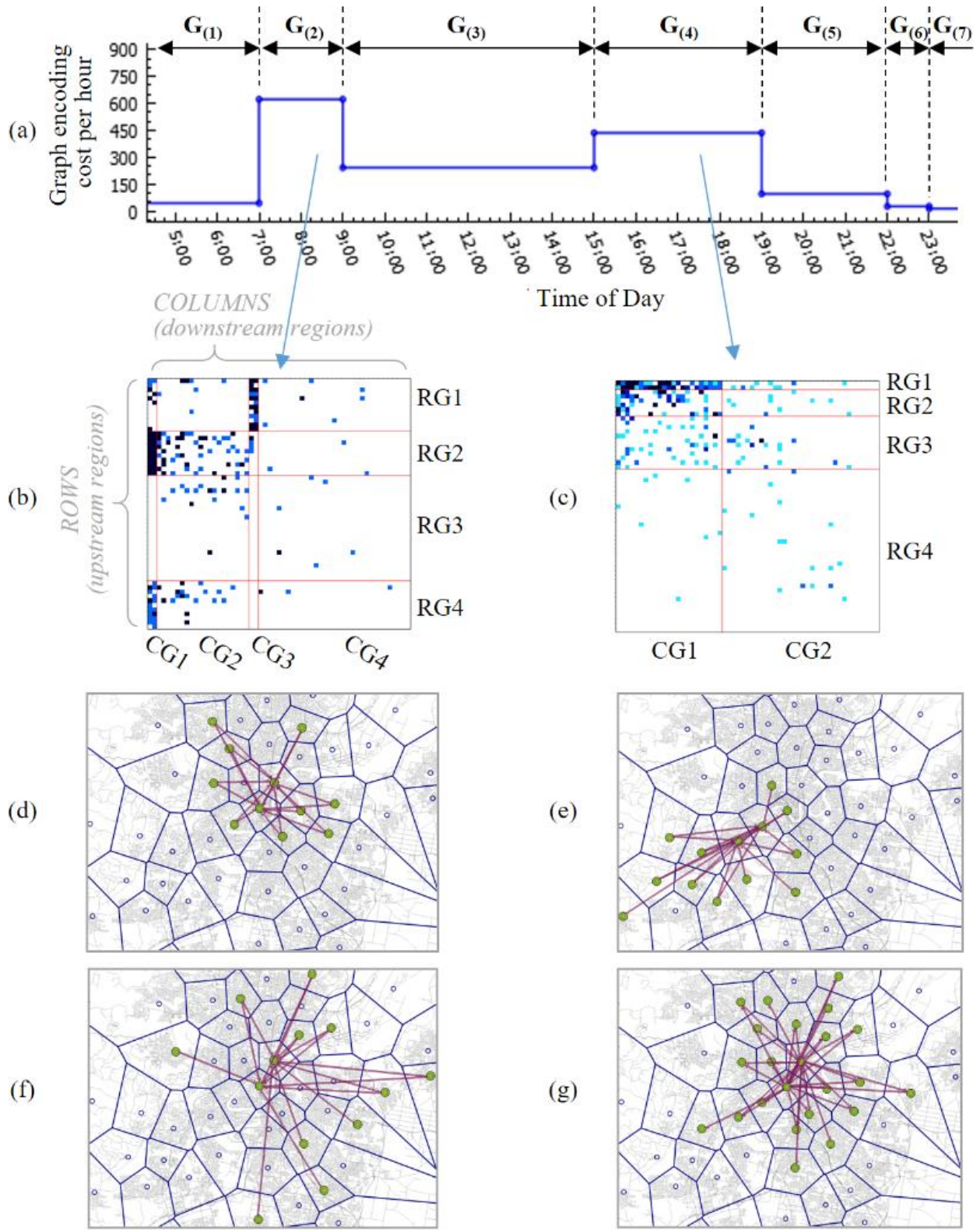

**FIGURE 5 Change point detection results for a selected graph series.**

**CONCLUSION**

This paper proposes a graph-based analysis framework that aims to characterise spatial and temporal patterns of network-wide traffic flows. We propose the use of a *trajectory flow graph* (or *trajectory flow map*), a dynamic graph of aggregated flows constructed from individual trajectories, to better understand and analyse city-wide mobility patterns. The proposed framework aims to apply data mining and graph mining techniques, coupled with a rich set of real-world trajectory data, to provide new ways of understanding, analysing, and visualizing traffic dynamics in a city. The paper presents approaches to partitioning the entire network into spatial cells and discusses methods to construct a trajectory flow graph at each time slice. The study applies a change point detection technique to divide the entire graph series into a set of graph. Using real-world trajectory data, this paper demonstrates the steps of spatial partitioning (cell construction) and temporal segmentation (change-point detection). The graph-representation obtained from the case study shows a concise yet meaningful description of time-evolving patterns in the underlying trajectory data and allows us to create a robust profile of network-wide traffic flows that can be used in various applications such as anomaly detection and congestion prediction.